\documentclass[runningheads]{llncs}
\usepackage[T1]{fontenc}
\usepackage{graphicx}
\usepackage{todonotes}
\usepackage{amsmath}

\begin{document}
\title{
Exact Certification of Data-Poisoning Attacks Using Mixed-Integer Programming
}
\titlerunning{Complete Verification of Data Poisoning Attacks}
\author{Philip Sosnin\inst{1}\orcidID{0009-0006-8929-4624} \and Jodie Knapp\inst{2} \and Fraser Kennedy\inst{2} \and Josh Collyer\inst{2}  \and Calvin Tsay\inst{1}\orcidID{0000-0003-2848-2809}}
\authorrunning{P. Sosnin et al.}
\institute{Department of Computing, Imperial College London, United Kingdom \email{\{p.sosnin23, c.tsay\}@imperial.ac.uk} \and The Alan Turing Institute, United Kingdom}
\maketitle        
\begin{abstract}
This work introduces a verification framework that provides both sound and complete guarantees for data poisoning attacks during neural network training.
We formulate adversarial data manipulation, model training, and test-time evaluation in a single mixed-integer quadratic programming (MIQCP) problem.
Finding the global optimum of the proposed formulation provably yields  worst-case poisoning attacks, while simultaneously bounding the effectiveness of all possible attacks on the given training pipeline. 
Our framework encodes both the gradient-based training dynamics and model evaluation at test time, enabling the first exact certification of training-time robustness.
Experimental evaluation on small models confirms that our approach delivers a complete characterization of robustness against data poisoning.
\keywords{Data Poisoning  \and Verification \and Mixed-integer Programming.}
\end{abstract}

\section{Introduction}

Data poisoning attacks pose a fundamental threat to the integrity of machine learning systems.
By injecting maliciously crafted samples into training data, adversaries can manipulate model behavior in ways that persist through deployment \cite{carlini2023poisoning,tian2022comprehensive}.
Recent work has shown that maliciously crafted attacks affecting less than 1\% of training samples can severely compromise model predictions \cite{han2022physical,zhu2019transferable}.
These vulnerabilities create an urgent need for formal verification methods that provide provable guarantees about training-time robustness.

While the study of inference-time robustness has led to a mature field of certified defences and verification algorithms \cite{dvijotham2018dual,fischetti2018deep,huchette2023deep,singh2019,wong2018scaling,zhang2018efficient}, the problem of certifying robustness to training-time perturbations remains comparatively under-explored. 
In contrast to test-time attacks, where adversarial inputs are evaluated on a fixed model, poisoning attacks alter the learning process itself, potentially influencing every parameter of the trained model and every subsequent prediction. 
This fundamental coupling between data, optimization, and model behavior makes certification of training-time robustness uniquely challenging.
Several recent efforts have introduced alternative approaches to certifying training-time robustness through modified model architectures or inference procedures.
For example, aggregation-based defences \cite{levine2020deep,rezaei2023run,wang2022improved} provide robustness guarantees for large ensembles of models, and randomized smoothing at inference time can provide guarantees for linear and kernel models \cite{rosenfeld2020certified}.
While effective in certain limited settings, these methods rely on strong assumptions about model structure and do not extend naturally to generic training settings.

To overcome these limitations, our previous work introduces a certification framework capable of reasoning about arbitrary training pipelines.
In particular, our Abstract Gradient Training (AGT) \cite{sosnin2024certified,sosnin2025abstractgradienttrainingunified,wicker2024certification} framework uses relaxations of the training dynamics that propagate bounds through gradient updates.
More recently, MIBP-CERT \cite{lorenz2024bicert,lorenz2024fullcert} takes an optimization-centered approach to these bounds.
These methods formulate training-time certification as a reachability analysis problem, applying interval or optimization-based relaxations at each training iteration to over-approximate the effect of data perturbations on the final learned parameters.
Such relaxations allow for computationally scalable certification, and enable general-purpose frameworks for bounding the effect of training-time perturbations without modifying the training algorithm itself.

A key challenge, however, lies in the trade-off between \textit{soundness} and \textit{completeness}.
A certification algorithm is \textit{sound} if every guarantee it produces is valid, i.e., no false assurances are made, but may be \textit{incomplete} if it cannot capture all possible safe behaviors.
Sound but incomplete methods, such as AGT and MIBP-CERT, rely on relaxations that over-approximate the reachable set of parameters at each training iteration.
While this improves the computational scalability to large models, it can lead to vacuous or overly conservative bounds.

In this work, we close this gap by introducing the first {sound and complete verification algorithm} for robustness to training-time attacks.
Our key contribution is a novel formulation of the joint training–attack–evaluation process as a single \textit{mixed-integer quadratic constrained program (MIQCP)}, taking advantage of the sound-and-complete nature of mixed-integer programming solvers.
Our formulation exactly encodes the interactions between adversarial data manipulations, gradient-based optimization steps, and test-time objectives, allowing us to reason about a \textit{given training procedure} (i.e., fixed initialization and data ordering) within a single optimization problem. 
By solving this MIQCP to optimality, we can compute both provably optimal poisoning attacks and exact robustness guarantees for any model trained under the specified threat model.
\\[0.5em]
\noindent\textbf{Contributions.} Our main contributions are summarised as follows:
\begin{itemize}
    \item We introduce the first {sound and complete verification framework} for data poisoning attacks on gradient-trained models, formulated as a MIQCP.
    \item We develop several tailored MIQCP approaches, including reformulations, heuristics, and bound tightening approaches for the proposed framework.
    \item We demonstrate, through empirical evaluation, that our approach achieves exact certification for small, linear models.
\end{itemize}

While computationally expensive compared to incomplete methods, our proof-of-concept framework can help reveal the structure of optimal poisoning attacks, identify where relaxations in existing certification tools introduce looseness, and inspire new certification and attack algorithms based on MIQCP.

The remainder of this paper is structured as follows. 
Section~\ref{sec:related_works} reviews related work on poisoning attacks and certification. 
Section~\ref{sec:background} formalizes our problem setting.
Sections~\ref{sec:miqp_formulation}--\ref{sec:tight_formulation} introduce our MIQCP-based verification framework and various improved solution strategies.
Finally, Section~\ref{sec:experiments} presents empirical results.

\section{Related Works}
\label{sec:related_works}

\subsection{Inference-Time Adversarial Attacks} 
\label{subsec:inference_verification}

Deep neural networks are highly vulnerable to small, carefully designed input perturbations that cause misclassification while remaining nearly imperceptible to humans \cite{szegedy2013intriguing,goodfellow2014explaining}. 
Such \textit{adversarial attacks} have motivated \textit{formal verification} methods that provide provable robustness guarantees within bounded perturbation regions \cite{gowal2018effectiveness,wong2018scaling,zhang2018efficient}. 
Formally, given a trained machine learning model $f_\theta$ and a property $\phi(x, f_\theta(x))$, verification aims to determine whether $\phi$ holds for all $x$ in some set $\mathcal{X}$.
An algorithm is \textit{sound} if it never certifies a false property and \textit{complete} if it can always identify $\phi$ holds when true.

\textbf{Incomplete Methods.}
Incomplete approaches, often referred to as \textit{certification} methods, ensure soundness by over-approximating the possible outputs of a network under the given set of input perturbations.
Representative techniques include interval bound propagation \cite{gowal2018effectiveness}, convex relaxations \cite{weng2018towards,zhang2018efficient}, and abstract interpretation frameworks \cite{singh2019}, which can scale to large architectures and be integrated into certified training pipelines \cite{muller2022certified,wong2018provable}.
However, these over-approximations are typically conservative, often yielding vacuous guarantees even when the model is robust.

\textbf{Complete Methods.}
Complete verification algorithms compute \textit{exact} robustness guarantees by exhaustively reasoning over all possible inputs and activation patterns. 
These methods typically encode the network and property as a constrained optimization problem using mixed-integer programming (MIP) or satisfiability modulo theory (SMT) solvers \cite{bunel2018unified,fischetti2018deep,katz2017reluplex,huchette2023deep,tsay2021partition}. 
Representative examples include ReLUplex \cite{katz2017reluplex}, Planet \cite{ehlers_formal_2017}, and NeuralSAT \cite{duong2023dpll}. 
These approaches are both sound and complete, but can incur high computational costs \cite{anderson2020strong,sosnin2024scaling,tjeng2017evaluating}. 
Our work builds directly on this line of complete methods, extending MIP from inference-time robustness to verification over training-time perturbations.

\subsection{Adversarial Data Poisoning Attacks}
\label{subsec:poisoning_certification}

Data poisoning attacks manipulate data \textit{during training} to degrade performance or induce targeted behaviors \cite{biggio2014poisoning,biggio2018wild,newsome2006paragraph}. 
Attacks are typically categorized as \textit{untargeted} (reduce model performance), \textit{targeted} (misclassify specific inputs), or \textit{backdoor} \cite{tian2022comprehensive}.
Backdoor attacks introduce hidden triggers that cause misclassification only when specific patterns are present \cite{chen2017targeted,gu2017badnets,han2022physical}, and even small fractions of poisoned data can cause severe failures \cite{yang2017fake,zhu2019transferable}.
Adversaries may be bounded or unbounded and are often assumed to have full knowledge of the model and training process.  
Defending against poisoning during training is challenging, as traditional strategies are often attack- or model-specific. 
These include training classifiers to detect poisoned inputs \cite{li2020learning}, applying noise or clipping to limit perturbations \cite{hong2020effectiveness}, or combining multiple models via disjoint dataset partitioning \cite{levine2020deep,rezaei2023run,wang2022improved}. 
While effective in specific scenarios, these approaches either fail to generalize to novel attacks or do not apply to general training settings.

On the other hand, certified robustness methods aim to provide guarantees against general poisoning strategies. 
For linear models, bounds have been established for gradient-based and $\ell_2$ perturbation attacks \cite{rosenfeld2020certified,steinhardt2017certified}, and differential privacy can provide statistical guarantees in limited settings \cite{xie2022uncovering}. 
More recent methods extend these guarantees to neural networks using interval bound propagation and MIP \cite{lorenz2024bicert,lorenz2024fullcert,sosnin2024certified,sosnin2025abstractgradienttrainingunified}, though the relaxations used result in incomplete certificates.
The work of \cite{sabanayagam2024exact} computes sound and complete certificates for label poisoning in graph neural networks, but with respect to a Neural Tangent Kernel linearisation, which is only exact in the infinite-width limit; for finite-width networks this approximation may yield incomplete guarantees.

Our approach provides the first general framework for sound and complete certification of standard, unmodified training procedures under a broad class of poisoning threats described in Section~\ref{subsec:threat_models}.
We explicitly state the assumptions required for our guarantees and their implications in Section~\ref{subsec:certification_problem}.

\section{Background}
\label{sec:background}

We first review the necessary background on supervised neural network training and formal methods relevant to poisoning robustness.
In particular, we consider a supervised learning setting in which a model $f(x, \theta)$ with parameters $\theta$ maps inputs $x$ from a feature space to outputs $y$ in a target space, which may be discrete, continuous, and/or multivariate. 
The model parameters $\theta$ are trained using a labelled dataset $\mathcal{D} = \{ (x^{(i)}, y^{(i)}) \}_{i=1}^N$ of $N$ input-output pairs.

\subsection{Neural Networks}  
While our framework is general, we focus our exposition on feed-forward neural networks with ReLU activation functions.
A feed-forward network with $K$ layers is defined by parameters $\theta = \{(W_k, b_k)\}_{k=1}^K$ and the layer-wise transformations
\begin{align}
    u_k &= W_k z_{k-1} + b_k, \qquad
    z_k = \sigma(u_k), \quad \forall k \in {1,...,K},
\end{align}
with input $z_0 = x$, output $f(x, \theta) = u_K$, and activation function $\sigma$, which we take to be ReLU.

Model parameters are typically learned via gradient-based optimization to minimize a loss function $\mathcal{L}$ that measures discrepancy between predictions and labels.
At each iteration $t$, SGD updates parameters on mini-batches $\mathcal{B}^{(t)} \subset \mathcal{D}$:
\begin{equation}
    \theta^{(t)} = \theta^{(t-1)} - \frac{\alpha^{(t)}}{|\mathcal{B}^{(t)}|} \sum_{(x^{(i)},y^{(i)}) \in \mathcal{B}^{(t)}} \nabla_\theta \mathcal{L}\big(f(x^{(i)},\theta^{(t-1)}),y^{(i)}\big),
\end{equation}
where $\alpha^{(t)} > 0$ is the learning rate at iteration $t$.  
We denote the parameters obtained by training $f$ on $\mathcal{D}$ (with some fixed training hyperparameters, parameter initialization, and data-ordering) as $\theta = M(\mathcal{D})$.

Although we restrict attention to vanilla SGD for clarity, other gradient-based training algorithms (e.g., momentum, Adam, or RMSProp) can be encoded within the same framework by introducing the corresponding auxiliary state variables and update equations, at the cost of increased formulation size.

\subsection{Data Poisoning Threat Models}
\label{subsec:threat_models}

We formalize poisoning adversaries through a \textit{perturbation model} $\mathcal{T}(\mathcal{D})$, which defines the set of possible modifications to the training dataset.  
We assume a white-box setting, where the adversary has full knowledge of the training data, model architecture, hyperparameters and parameter values at each training iteration.
This white-box setting encompasses dynamic adversaries who can adaptively choose their poisoning objectives during training based on the evolving model state~\cite{bose2025keeping}.
We consider poisoning attack capabilities in two broad classes:

\paragraph{1) Bounded Perturbations.}
In the bounded attack setting, an adversary can modify up to $n$ training samples within pre-defined constraints around the features and/or labels of the clean training samples.  
Formally, the perturbed dataset $\tilde{\mathcal{D}}=\{ (\tilde{x}^{(i)}, \tilde{y}^{(i)}) \}_{i=1}^N \in \mathcal{T}^{n,\epsilon,\nu}_{\text{bounded}}(\mathcal{D})$ satisfies
\begin{equation}
    \| x^{(i)} - \tilde{x}^{(i)} \|_p \le \epsilon, \quad
    \| y^{(i)} - \tilde{y}^{(i)} \|_q \le \nu, \quad
    \forall i \in \mathcal{I}, \quad |\mathcal{I}| \le n,
\end{equation}
where $\mathcal{I}$ indexes the modified samples.
We take the feature perturbation norm to be $p=\infty$, and the label perturbation norm to be $q=\infty$ and $q=0$ for regression and classification tasks, respectively.
Bounded perturbations include \textit{clean-label attacks} ($\nu=0$), where only features are modified, and \textit{label-flipping attacks} ($\epsilon=0$), where only labels are altered \cite{munoz2017towards,steinhardt2017certified,zhu2019transferable}.

\paragraph{2) Arbitrary (Unbounded) Perturbations.} 
In a more general threat model, an adversary may arbitrarily replace data points from the training set.
In this setting, the adversary can substitute up to $n$ training samples with entirely arbitrary points unrelated to the removed data, and may optionally modify their associated labels:
\begin{equation}
    \tilde{\mathcal{D}} = (\mathcal{D} \setminus \mathcal{S}) \cup \tilde{\mathcal{S}}, \quad |\mathcal{S}| = |\tilde{\mathcal{S}}| \le n,
\end{equation}
where $\mathcal{S}$ contains the removed samples and $\tilde{\mathcal{S}}$ contains the injected points.  
This formulation encompasses fully general poisoning attacks, with clean-label and label-flipping attacks as special cases when the injected points preserve or modify labels accordingly \cite{chen2017targeted,gu2017badnets,munoz2017towards}.  
To ensure that any resulting verification problem remains bounded, we assume the injected samples to lie within a bounded domain.
Unlike the bounded perturbation model above, these bounds are independent of the replaced data samples, and can be chosen to cover the entire input space if the domain is naturally constrained (e.g., physical measurements) or pre-clipped, effectively allowing arbitrary substitutions.

\remark{For both threat models, we assume the ordering (batching) of training samples remains fixed before and after poisoning. Specifically, each poisoned sample appears at the same position within mini-batches during training as its corresponding original sample, preserving the sequence in which data are presented across all training iterations.}

\paragraph{Adversarial Objectives.} We assume the adversary seeks to maximize an objective corresponding to a post-training measure of attack success:
\begin{itemize}
\item \textit{Untargeted attacks} aim to degrade overall model accuracy on a chosen dataset. A special case is the \textit{denial of service attack}, which aims to prevent model convergence on the training set.
\item \textit{Targeted attacks} aim to force specific outputs on selected inputs while leaving other predictions unaffected.
\item \textit{Backdoor attacks} aim to cause misclassification when a trigger is present, while keeping performance on clean inputs unchanged.
\end{itemize}
We denote the attack objective as a function $J(\theta)$ evaluated on the final trained model.
For instance, the test error on a dataset $\{x^{(i)}_{\text{test}}, y^{(i)}_{\text{test}}\}_{i=1}^{N{\text{test}}}$ can be written as 
$J(\theta) = \sum_{i=1}^{N_{\text{test}}} \mathbf{I}\left\{f^\theta(x^{(i)}_{\text{test}}) \neq y^{(i)}_{\text{test}}\right\}$.

\subsection{Verification Problem}
\label{subsec:certification_problem}

Our verification framework aims to certify that a machine learning model satisfies specified properties despite adversarial perturbations to its training data. We begin by establishing the assumptions under which our certificates are valid\footnote{We note that these assumptions match those adopted in \cite{lorenz2024bicert,lorenz2024fullcert,sosnin2024certified,sosnin2025abstractgradienttrainingunified}.}:
\begin{itemize}
    \item \textit{Assumption 1: Fixed Initialization and Data Ordering.} We assume fixed model parameter initialization and a fixed ordering of training samples. Consequently, our guarantees apply to a specific random seed controlling both initialization and data order during training.
    \item \textit{Assumption 2: White-Box Adversary with Constrained Influence.} We assume the adversary has complete knowledge of training hyperparameters, model architecture, parameter initialization, data ordering, and the training dataset. However, the adversary's influence is limited exclusively to the allowable perturbations of training data as defined by the perturbation model.
\end{itemize}
Let $\mathcal{D}$ denote the original dataset, $M$ the training procedure (including parameter initialization, data ordering, and hyperparameters), %
and $\mathcal{T}$ the perturbation model, as described above. The optimal adversarial manipulation can be written:
\begin{equation}
\max_{\tilde{\mathcal{D}}} J(\tilde{\theta}) \quad \text{s.t.} \quad \tilde{\theta} = M(\tilde{\mathcal{D}}), \quad \tilde{\mathcal{D}} \in \mathcal{T}(\mathcal{D}).
\end{equation}
Computing the global optimum of this problem captures the worst-case effect of allowable perturbations, with respect to $J(\cdot)$. Prior works \cite{lorenz2024bicert,sosnin2024certified,sosnin2025abstractgradienttrainingunified} compute provably valid upper bounds, rather than this worst case exactly.

\paragraph{Mixed-Integer Programming.}
We formulate this problem as a mixed-integer quadratically constrained program (MIQCP). 
Unlike interval or convex relaxation methods, MIQCP explicitly encodes discrete decisions and non-linear dependencies across the entire training procedure, allowing global optimization over all possible perturbations. This comes at a computational cost, as MIP is NP-Hard in general.  
Formally, a mixed-integer program involves continuous variables $a \in \bbbr^{n_c}$, integer variables $b \in \bbbz^{n_i}$, and constraints $g_j(a, b)$:
\begin{alignat}{2}
\min \ & h(a, b) \qquad \text{s.t. } g_j(a, b) \le 0, \quad j = 1, \dots, m,
\end{alignat}
where the $h$ and $g_j$ may include linear, quadratic, or bilinear terms. 

While \cite{lorenz2024bicert,sosnin2025abstractgradienttrainingunified} encode individual training iterations as MIPs, our approach encodes the \textit{entire training and testing procedure} in a single optimization problem, thus incurring no over-approximation. 

\section{A Mixed-Integer Formulation for Verification of Training-Time Attacks}
\label{sec:miqp_formulation}

Our formulation comprises three components:  
(1) \emph{data-perturbation constraints}, defining the allowed training-set manipulations;  
(2) \emph{training-dynamics constraints}, encoding gradient computation and parameter updates; and  
(3) \emph{test-time evaluation constraints}, encoding the adversarial objective over the trained model.  

\subsection{Encoding Training Data Perturbations}

For each sample in $\{x^{(i)}, y^{(i)}\}_{i=1}^N$, we introduce perturbed variables $(\tilde{x}^{(i)}, \tilde{y}^{(i)})$ and a binary indicator  
$s^{(i)} \in \{0,1\}$
denoting whether sample $i$ is modified. 
Here we provide formulations for both threat models described in Section~\ref{subsec:threat_models} assuming $y^{(i)} \in \{0, 1\}$, but these may be extended to multi-class and regression settings.
The adversarial budget $n$ is enforced by the constraints $\sum_i s^{(i)} \le n$.

\paragraph{Bounded Feature and Label Perturbations.}
Perturbations under the bounded threat model can be defined using the following constraints:
\begin{equation}
\left.
\begin{aligned}
& x^{(i)} - \varepsilon\, s^{(i)} \;\le\; \tilde{x}^{(i)} \;\le\; x^{(i)} + \varepsilon\, s^{(i)}, \\
& y^{(i)} (1 - s^{(i)}) \;\le\; \tilde{y}^{(i)} \;\le\; y^{(i)} (1 - s^{(i)}) + s^{(i)}, \quad \\
& \tilde{x}^{(i)} \in \bbbr^{d},\quad \tilde{y}^{(i)},s^{(i)} \in \{0,1\},
\end{aligned}
\right\}
\quad  i=1,\dots,N.
\end{equation}

\paragraph{Arbitrary Substitutions.}
For the arbitrary substitution threat model, each sample may be replaced by any point drawn from some MIQCP-representable domain $\mathcal{X}^a$. 
Using big-$M$ constants $L^x, U^x \in \bbbr^{d}$, which satisfy $L^x_j \leq x_j \leq U^x_j$ for all $x \in \mathcal{X}^a$, we encode:
\begin{equation}
\left.
\begin{aligned}
& x^{(i)}(1 - s^{(i)}) + L^x\, s^{(i)} \;\le\; \tilde{x}^{(i)} \;\le\; x^{(i)}(1 - s^{(i)}) + U^x\, s^{(i)}, \ \\
& y^{(i)}(1-s^{(i)}) \;\le\; \tilde{y}^{(i)} \;\le\; y^{(i)}(1-s^{(i)}) + s^{(i)},  \\
& \tilde{x}^{(i)} \in \mathcal{X}^a,\quad \tilde{y}^{(i)},s^{(i)} \in \{0,1\},
\end{aligned}
\right\}
\  i=1,\dots,N.
\end{equation}

\subsection{Encoding the Model Training Dynamics}
\label{subsec:training_dynamics}

We denote weight matrices $W_k^{(t)}$, biases $b_k^{(t)}$, pre-activations $u_k^{(t)}$, and activations $z_k^{(t)}$ at iteration $t$ for layers $k=1,\dots,K$.
Then, denoting $\mathcal{I}^{(t)}$ as the index set that specifies samples included in the SGD batch according to the fixed data ordering, $z^{(t, i)}_0 = \tilde{x}^{(i)}, i \in \mathcal{I}^{(t)}$ are the inputs considered at iteration $t$.

\paragraph{Forward Pass.}
At each layer $k$ of the neural network, the pre-activation vector is given by the bilinear constraint ($W_k$ are variable cf. inference-time verification):
\begin{equation}\label{eq:bilinear}
u_k^{(t, i)} = W_k^{(t)} z_{k-1}^{(t, i)} + b_k^{(t)}.
\end{equation}
To encode the ReLU activations $z_k^{(t, i)} = \max\{0, u_k^{(t, i)}\}$, we introduce activation binaries
$a_k^{(t, i)} \in \{0,1\}^{n_k}$.
For each neuron in the network, we require big-$M$ constants $L^{(t, i, \mathrm{relu})}_{k}, U^{(t, i, \mathrm{relu})}_{k}$ satisfying $L^{(t, i, \mathrm{relu})}_{k} \leq u_k^{(t, i)} \leq U^{(t, i, \mathrm{relu})}_{k}$.
\footnote{The big-$M$ constants can be obtained using any sound bounding method, such as interval bound propagation.
Existing incomplete certifiers (e.g., \cite{sosnin2024certified}) naturally provide such bounds, which can be further tightened using bound tightening approaches.}
Then, the big-$M$ formulation of the ReLU activation function is encoded by the following:
\begin{equation}\label{eq:relubigm}
\left.
\begin{aligned}
&  u_k^{(t, i)} \le z_k^{(t, i)} \le u_k^{(t, i)} - L^{(t, i, \mathrm{relu})}_{k} \odot \left( \mathbf{1} - a_k^{(t, i)} \right), \quad   \\
& 0 \le z_k^{(t, i)} \le U^{(t, i, \mathrm{relu})}_{k} \odot a_k^{(t, i)}. \\
\end{aligned}
\right\}
\quad 
\begin{aligned}
i&\in \mathcal{I}^{(t)},\\
k&=1,\dots,K,\\
t&=0,\dots,T.
\end{aligned}
\end{equation}
Here we use $\odot$ to denote element-wise multiplication.
When $a_{k,j}^{(t, i)}=1$, the $j$th neuron in layer $k$ is active and $z_{k,j}^{(t, i)} = u_{k,j}^{(t, i)}$;  
when $a_{k,j}^{(t, i)}=0$ the unit is inactive and $z_{k,j}^{(t, i)} = 0$.  
Overall, \eqref{eq:bilinear} is quadratic and \eqref{eq:relubigm} is mixed-integer linear.

\paragraph{Training Loss.}
We select the hinge loss $\mathcal{L}(\hat{y}, y) = \operatorname{ReLU}\left(1 - (2 y - 1) \hat{y}\right)$ as our training loss function, since it is readily MIP-representable (we use $2y - 1$ to transform the labels into the domain $\{-1, 1\}$).
For training sample $i$ at iteration $t$ with predicted logit $\hat{y}^{(t, i)} = u^{(K, i)} \in \bbbr$, we define the auxiliary variables $r^{(t, i)} = 1 - ( 2 \tilde{y}^{(i)} - 1) \hat{y}^{(t, i)}.$
We introduce hinge-activation binary variables $h^{(t, i)}\in\{0,1\}$ and big-$M$ constants $L^{(t, i, \mathrm{hinge})}, U^{(t, i, \mathrm{hinge})}$ that bound $r^{(t, i)}$. Then, the $\operatorname{ReLU}$ constraint can be represented using the same big-$M$ formulation described in~\eqref{eq:relubigm}: 
\begin{equation}
\left.
\begin{aligned}
& r^{(t,i)} \le \mathcal{L}^{(t,i)} \le r^{(t,i)} - L^{(t, i, \mathrm{hinge})} \left( 1 - h^{(t, i)} \right), \quad  \\
& 0 \le \mathcal{L}^{(t,i)} \le U^{(t, i, \mathrm{hinge})} h^{(t, i)}. \\
\end{aligned}
\right\}
\quad \begin{aligned}
i& \in \mathcal{I}^{(t)},\\ t&=0,\dots,T.  
\end{aligned}
\end{equation}
The derivative of the loss with respect to each predicted logit can be defined using the bilinear constraints
\begin{align}
\frac{\partial \mathcal{L}^{(t,i)}}{\partial \hat{y}^{(t,i)}} = -2\tilde{y}^{(i)} h^{(t,i)} .
\end{align}
The equivalent training loss constraints for regression settings with the {squared error loss} are simply $\mathcal{L}^{(t, i)} = (\hat{y}^{(t, i)} - \tilde{y}^{(i)})^2$ and ${\partial \mathcal{L}^{(t, i)}}/{\partial \hat{y}^{(t, i)}} = 2 (\hat{y}^{(t, i)} - \tilde{y}^{(i)})$ for all $i \in \mathcal{I}^{(t)}$ and $t=0,\dots,T.$ Unlike the classification case, these constraints do not require auxiliary binary variables or the big-$M$ formulation.

\paragraph{Backward Pass.}
The constraints for the backward pass follow the standard backpropagation rules, which can be written as linear and bilinear constraints.
The activation patterns of the $\operatorname{ReLU}$ operations are already captured through the binary variables in \eqref{eq:relubigm}, which can be reused in the backward pass.
Again letting $\odot$ denote element-wise multiplication, the gradients for layer $k$ satisfy:
\begin{equation}
\left.
\begin{aligned}
\frac{\partial \mathcal{L}^{(t,i)}}{\partial u_k^{(t, i)}} &= 
    \frac{\partial \mathcal{L}^{(t,i)}}{\partial z_k^{(t, i)}} \odot a_k^{(t, i)}, \ 
\frac{\partial \mathcal{L}^{(t,i)}}{\partial z_{k-1}^{(t, i)}} = 
    \left(W_k^{(t)}\right)^\top \frac{\partial \mathcal{L}^{(t,i)}}{\partial u_k^{(t,i)}}, \ \\
\frac{\partial \mathcal{L}^{(t,i)}}{\partial W_k^{(t)}} &=
    \frac{\partial \mathcal{L}^{(t,i)}}{\partial u_k^{(t,i)}} \, \left(z_{\,k-1}^{(t,i)}\right)^\top, \ 
\frac{\partial \mathcal{L}^{(t,i)}}{\partial b_k^{(t)}} = \sum_{i\in\mathcal{I}^{(t)}}
    \frac{\partial \mathcal{L}^{(t,i)}}{\partial u_k^{(t,i)}} .
\end{aligned}
\right\}
\  k=1,\dots,K.
\end{equation}

\paragraph{Parameter Updates.}
For learning rates $\alpha^{(t)}$ and mini-batches indexed by $\mathcal{I}^{(t)}$, the parameter updates are encoded by the linear constraint
\begin{equation}
\theta^{(t)} = \theta^{(t-1)} 
  - \frac{\alpha}{|\mathcal{I}^{(t)}|}
    \sum_{i\in \mathcal{I}^{(t)}} \delta^{(t,i)},
\qquad t=1,\dots,T,
\end{equation}
where $\delta^{(t,i)}$ stacks all layerwise parameter gradients.

\subsection{Encoding Post-Training Evaluation}

We now formulate the adversarial attack goal as the objective function of our optimization problem.
We focus on untargeted attacks, giving examples for attacks that degrade test-time performance or cause denial of service.
Extensions to targeted/backdoor objectives follow the same pattern and are discussed below.

\paragraph{Degrading Model Accuracy.}
First, we present the constraints required to represent an attacker that wishes to maximize the classification error on a particular test set
$\{(x_{\mathrm{test}}^{(i)}, y_{\mathrm{test}}^{(i)})\}_{i=1}^{N_{\mathrm{test}}}$.
For each test sample $i$, we encode a forward pass through the network using the final weights
$W_k^{(T)}, b_k^{(T)}$.
Let $z_{K}^{(i, \text{test})}$ denote the final-layer logits for test example $i$ obtained
via the forward-pass constraints \eqref{eq:bilinear}--\eqref{eq:relubigm}.
The model's binary prediction is then
\begin{align}
p^{(i, \text{test})} = \begin{cases}
    1 &\quad \text{if } z_{K}^{(i, \text{test})} \geq 0,\\
    0, & \quad \text{otherwise.}
\end{cases}
\end{align}
We introduce one binary variable $p^{(i, \text{test})}\in\{0,1\}$ into our MIQCP formulation per test example to encode this model prediction.
Using big-$M$ constants $L^{(i, \mathrm{test})}, U^{(i, \mathrm{test})}$,
the predicted label is encoded by:
\begin{align}
L^{(i, \mathrm{test})}\left(1 - p^{(i, \text{test})}\right) \leq z_K^{(i, \text{test})} \le U^{(i, \mathrm{test})}\, p^{(i, \text{test})} - \epsilon \left(1 - p^{(i, \text{test})}\right)
\end{align}
where $\epsilon > 0$ is any small constant used to break ties at 0.
Thus $p^{(i, \text{test})} = 1$ implies a class $1$ prediction, while
$p^{(i, \text{test})} = 0$ corresponds to class $0$.
Finally, we can define the total test error of the model to be
\begin{align}
J\left(\theta^{(T)}\right) = \sum_{i: y_{\mathrm{test}}^{(i)} = 1} p^{(i, \text{test})} + \sum_{i: y_{\mathrm{test}}^{(i)} = 0} 1 - p^{(i, \text{test})}.
\end{align}

\paragraph{Denial of Service.} 
A second attack objective seeks to degrade training convergence, thereby rendering the trained model ineffective.
We encode this goal using the training loss variables $\mathcal{L}^{(t,i)}$ defined earlier.
While an attacker could target specific training iterations (e.g., losses in the final epoch), here we formulate the objective as maximizing the cumulative training loss given by $\sum_{t=1}^{T} \sum_{i\in \mathcal{T}^{(t)}}\mathcal{L}^{(t,i)}$.
This causes the optimization to choose poisoning samples that impair the learning process across all training steps, rather than just the final iteration.

\paragraph{Alternative Attack Goals.}
The formulations above extend naturally to targeted and backdoor attack objectives.  
For a targeted attack, instead of penalizing disagreement with the true label, the
binary prediction variables are encouraged to match a specified target
label for each test point, and the objective maximizes the number of induced target
classifications.
For a backdoor attack, certificates can be obtained by additionally permitting
adversarial manipulation of the test-time inputs themselves, thereby encoding the
possibility of injected trigger patterns when evaluating the model’s predictions.

\remark{
In our complete MIQCP formulation, any feasible primal solution corresponds to a valid poisoning attack, while any dual bound provides a sound certificate on the worst-case poisoning.  
Both the optimal attack and the exact certificate are obtained by solving the MIQCP.
Consequently, existing data poisoning attacks can supply primal solutions to the MIP solver, and, conversely, primal solutions found by the solver directly constitute strong poisoning attacks.
Similarly, tighter relaxations improve the resulting certificates given by the dual bound.  
Section~\ref{sec:primal} demonstrates poisoning heuristics for improving primal solutions, while Sections~\ref{sec:obbt}--\ref{sec:auxiliary} describe methods for tightening the formulation.
}

\section{Improved Formulations and Solution Strategies}
\label{sec:tight_formulation}

MIQCPs are generally solved using a branch-and-bound algorithm, where computational efficiency depends on two primary factors: the tightness of the continuous relaxations, and the quality of primal feasible solutions used in pruning and node selection strategies.
This section introduces strategies to address these challenges, yielding faster and more reliable certification performance.

\subsection{Local Search Primal Heuristic}\label{sec:primal}

Standard branch-and-bound solvers treat all decision variables as equally free to vary subject to their constraints, without exploiting any problem-specific structure. In our training-time certification problem, however, the entire objective and all relevant variables are ultimately a deterministic function of the manipulated training data. In other words, significant effort can be spent exploring parameter variables whose values are entirely specified once the label or feature perturbations are fixed. By focusing the search on feasible perturbations, a primal heuristic can more efficiently explore the space of possible solutions.

To exploit this problem structure, we develop a specialized primal heuristic that performs a local search over the poisoning assignment binaries $s^{(i)}$ using fast, batched retraining of the model.
The heuristic starts from the current incumbent provided by the MIP solver and explores a sequence of Hamming-ball neighborhoods around this center.
At each search radius, it constructs candidate poisoning assignments by swapping a small number of poisoned and clean labels, while preserving the adversary’s budget.
Each batch of candidates is then evaluated in parallel by retraining under a batched GPU-based SGD procedure, yielding their corresponding test losses.
Whenever a candidate improves upon the current incumbent solution, the heuristic immediately returns the updated poisoning assignment to the solver; if no improvement is found, the search radius expands until either a new incumbent arrives from the solver or all feasible neighborhoods have been exhausted.

In certain cases (namely, label-flipping attacks), the primal heuristic can prove optimality by exhaustive search when the search radius exceeds the total poisoning budget. 
In these instances, the best candidate returned by the heuristic is certifiably optimal.
Due to the combinatorial growth of the search space, this situation arises only for very small budgets, e.g., $n \leq 4$.
Future work may explore incorporating existing poisoning attacks to further improve primal heuristics.

\subsection{Bounds Tightening within the Test Data Hull}
\label{sec:obbt}

The strength of the continuous relaxation in a Mixed-Integer Quadratic Program (MIQCP) depends heavily on the initial bounds of continuous variables and the resulting big-$M$ constants.
To improve these bounds before optimization, \emph{Optimization-Based Bounds Tightening} (OBBT) is commonly applied \cite{gleixner2017three,sosnin2024scaling}.
OBBT refines variable domains by solving auxiliary optimization problems that yield tighter upper and lower bounds.

In classification settings, the constraints with largest effect on the continuous relaxation are those defining the test predictions $p^{(i,\text{test})}$, which rely on big-$M$ constants $L^{(i,\text{test})}$ and $U^{(i,\text{test})}$ that bound the final-layer logits $\hat{y}^{(t,i)}$.
These bounds accumulate looseness from all preceding layers, and tightening them is crucial for obtaining a strong dual bound.

Directly tightening the logit bounds for all $N^{\text{test}}$ test samples requires solving $2N^{\text{test}}$ MIQCPs, which is often impractical because each has complexity comparable to the original problem.
Shorter time limits produce valid, but looser big-$M$ constants.
To balance this trade-off, we introduce a single auxiliary input variable $\boldsymbol{x}^{\text{aux}}$ constrained to lie in the integer hull of the test set:
\begin{equation}
\boldsymbol{x}^{\text{aux}} \in \operatorname{IntHull}\!\left\{\boldsymbol{x}^{(1,\text{test})},\dots,\boldsymbol{x}^{(N^{\text{test}},\text{test})}\right\}.
\end{equation}
We then encode its forward pass to obtain $\hat{y}^{\text{aux}}$ and solve two bounding problems (min/max). The resulting values $L^{\text{aux}}$ and $U^{\text{aux}}$ jointly bound the logits of \emph{all} test samples. This reduces the cost to two sub-problems, allowing longer sub-problem solve times and yielding tighter big-$M$ constants.

This idea extends naturally by clustering test points into $P$ sub-groups (e.g., by class) and constructing two auxiliary sub-problems per group, solving the resulting $2P$ problems trades additional computation for tighter per-group bounds.

\subsection{Auxiliary Variable Formulation}
\label{sec:auxiliary}

The arbitrary substitution threat model is the most expressive, but also the most computationally challenging.
When the decision variables $\tilde{x}^{(i)}, ..., \tilde{x}^{(N)}$ corresponding to each training sample may lie anywhere in the attacker domain $\mathcal{X}^a$, the resulting big-$M$ constants for all forward and backward passes must accommodate worst-case values over $\mathcal{X}^a$.
This substantially weakens the continuous relaxation, even for samples that are  unmodified in the optimal solution.

To mitigate this effect, we introduce an auxiliary variable formulation that decouples clean and adversarial data at the level of network propagation and gradient computation.
The idea is to ensure that only adversarially substituted samples incur loose bounds, while clean training points retain tight bounds derived from the original dataset.

Specifically, we introduce decision variables for two classes of training data:
\begin{enumerate}
\item \textit{Clean data.}
Each original sample $(x^{(i)}, y^{(i)})$ is propagated through the network using the standard forward and backward constraints described in Section~\ref{subsec:training_dynamics}.
Since these samples are not drawn from $\mathcal{X}^a$, their activations admit significantly tighter bounds, yielding a stronger relaxation.
Let the resulting gradient decision variables for these samples be denoted $\delta^{(t, i)}$.
\item \textit{Auxiliary data.}  
We introduce $n$ auxiliary samples $(\tilde{x}^{(j)}, \tilde{y}^{(j)})$, $j=1,\dots,n$, representing points that will be substituted into the dataset. These variables satisfy
$
    \tilde{x}^{(j)} \in \mathcal{X}^a, 
    \tilde{y}_{\mathrm{aux}}^{(j)} \in \{0,1\},
$
and are propagated independently through the network's forward and backward passes, resulting in (perturbed) gradient decision variables $\tilde{\delta}^{(t, j)}$.
\end{enumerate}

Rather than model manipulations directly in the input space, we now enforce substitutions at the gradient level.
We first introduce the following binary decision variables:
\begin{itemize}
    \item $s^{(i)} \in \{0,1\}$ denotes the removal of the original sample $x^{(i)}$.
    \item $\tilde{s}^{(i, j)} \in \{0,1\}$ denotes whether perturbed sample $\tilde{x}^{(j)}$ replaces $x^{(i)}$.
\end{itemize}
The poisoning adversary budget and unique substitutions are enforced via the constraint $\sum_{j=1}^n \tilde{s}^{(i, j)} = s^{(i)}$.
Then, to enforce the substitution of the data, we modify the parameter update constraint, now given by 
\begin{equation}
\theta^{(t)} = \theta^{(t-1)} -
\frac{\alpha^{(t)}}{\lvert \mathcal{I}^{(t)} \rvert}
\left[
\sum_{i \in \mathcal{I}^{(t)}} \left((1 - s^{(i)})\,\delta^{(t,i)}\;+\;\sum_{j=1}^{n} \tilde{s}^{(i, j)},\tilde{\delta}^{(t,j)}\right)
\right]\,.
\end{equation}
If an original point is retained, i.e., $s^{(i)}=0$, its tightly bounded gradient $\delta^{(t,i)}$ contributes to the update.
If it is removed, i.e., $s^{(i)}=1$, its contribution is suppressed and replaced by that of an activated auxiliary sample ($\tilde{s}^{(i, j)}=1$), with corresponding gradient $\tilde{\delta}^{(t,j)}$ to encode the worst-case substitution.

\begin{figure}[t]
    \centering
    \includegraphics{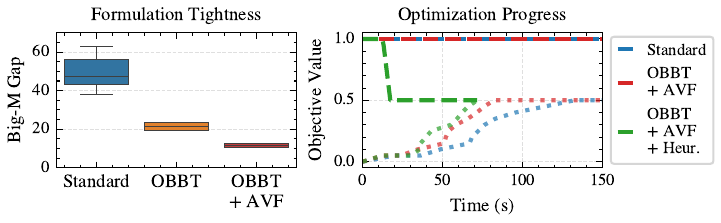}
    \vspace{-20pt}
    \caption{Comparison of formulation tightness and optimization progress for the Iris dataset under an unbounded attack model ($n=8$). Left: Tightness of the test-data Big-M constants, defined as $|U^{(i,\mathrm{test})} - L^{(i,\mathrm{test})}|$.
    Right: Objective value (dashed) and dual bound (dotted) progress over time.}
    \label{fig:iris_strategies}
    \vspace{-8pt}
\end{figure}

\section{Experimental Results}
\label{sec:experiments}

We demonstrate our framework on several moderate-scale learning problems where exact verification is computationally feasible. 
Note that, exact certification/optimal attacks are often intractable for larger scale problems and impose similar difficulty for MIQCP (see Remark 2). 
The goal of these experiments is to (i) empirically validate the proposed MIQCP formulation, (ii) qualitatively and quantitatively assess optimal poisoning attacks under different threat models, and (iii) study the effectiveness of the tightening strategies in Section \ref{sec:tight_formulation}.

All experiments were run on a server equipped with an AMD EPYC 9334 CPU using Gurobi~13.0.
We impose limits of 1 hour and 8 threads per instance; unless explicitly stated, we report optimal attacks only where the solver has proven global optimality (the default MIP gap is reached).
\\[0.5em]
\noindent\textbf{Datasets.}
We evaluate our method on three benchmark datasets:
\begin{itemize}
    \item {Iris} ($d=4$, $N=80$, $N^{\mathrm{test}}=20$):  
    The classical Iris dataset~\cite{anderson1936species}, restricted to a binary classification task by selecting the first two classes.
    \item {Diabetes} ($d=10$, $N=320$, $N^{\mathrm{test}}=89$):  
    The diabetes dataset introduced by \cite{efron2004least}, in which the regression task is to predict a quantitative measure of disease progression.
    \item {Halfmoons} ($d=2$, $N=100$, $N^{\mathrm{test}}=40$):  
    A synthetic two-dimensional binary classification dataset with nonlinearly separable classes.
\end{itemize}

\noindent\textbf{Models.}
Across all experiments, we use a single linear layer as the learning model.
In classification settings, this corresponds to a linear support vector machine, while in regression settings it reduces to standard linear regression.
For the nonlinearly separable Halfmoons dataset, we apply a fixed polynomial feature expansion prior to training, resulting in a total of $9$ input features. Note that certification problems remain nonlinear with quadratic terms and loss functions. 

\begin{figure}[t]
    \centering
    \includegraphics{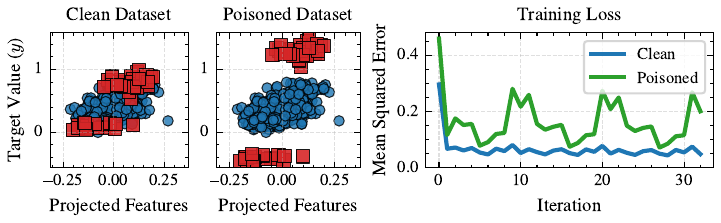}
    \vspace{-20pt}
    \caption{Optimal denial of service attack on the Diabetes dataset with $n=50, \epsilon=0, \nu=0.5$. The red squares depict the points poisoned by the adversary. The rightmost figure depicts the training loss for the poisoned (green) vs original (blue) datasets.}
    \label{fig:diabetes_poisoning}
    \vspace{-12pt}
\end{figure}

\paragraph{Effect of Solution Improvement Strategies.}
We evaluate the MIQCP solution improvement strategies introduced in Section~\ref{sec:tight_formulation} using an unbounded attack on the Iris dataset with $n=8$.
The model is trained for 4 epochs with a batchsize of 20 and a learning rate of $\alpha=0.03$.
Figure~\ref{fig:iris_strategies} illustrates the effect of our improvement strategies on both formulation tightness and solver runtime.

To quantify tightness, we measure the gap between the final Big-M constants on the test data, defined as $|U^{(i,\mathrm{test})} - L^{(i,\mathrm{test})}|$.
The left panel of Figure~\ref{fig:iris_strategies} compares this gap for the standard and tightened formulations.
As expected, OBBT significantly tightens the Big-M constants, and the auxiliary variable formulation yields additional tightening.
In terms of computational performance (Figure~\ref{fig:iris_strategies}, right), these tightening techniques accelerate convergence of the dual bound but have little effect on the primal solution. 
Intuitively, the tighter relaxations improve certification, but not identification of the optimal attack. 
By additionally incorporating our primal heuristic, we can rapidly identify strong attacks, leading to a substantial reduction in the time required to prove optimality.

\paragraph{Optimal Denial-of-Service Attack.}
We now evaluate a denial-of-service attack on a regression model trained on the Diabetes dataset.
We take the adversary's objective to be maximizing the average training loss throughout training, thus preventing model convergence.
Figure~\ref{fig:diabetes_poisoning} shows the resulting optimal attack for an adversary capable of modifying $n=50$ training samples by up to $\nu=0.5$ in the label space.
As seen in the training loss curve, our proposed framework reveals the adversary can effectively prevent stable model convergence by injecting perturbations that induce oscillatory behaviour in the training loss.

\begin{figure}[t]
    \centering
    \includegraphics{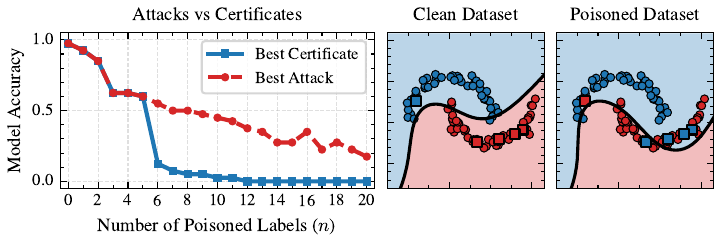}
    \vspace{-20pt}
    
    \caption{
    Right: Label flipping attacks and certificates for the Halfmoons dataset. Left: Example optimal attack for $n=5$ label flips.}
    \label{fig:halfmoons_boundaries}
\end{figure}

\paragraph{Optimal Label Flipping on the Halfmoons Dataset.}
Finally, we examine a poisoning attack on the Halfmoons dataset.
Figure~\ref{fig:halfmoons_boundaries} summarizes the attack's effectiveness.
On the right, we visually observe the certified worst-case impact of the attack on the model's decision boundary. 
The model trained on the poisoned dataset exhibits a significantly altered decision boundary, with relatively few label flips required to cause significant mis-classification in the resulting model.

The left side of Figure~\ref{fig:halfmoons_boundaries} shows the resulting optimal attacks and certified accuracies computed for various strengths of the label flipping attack.
We observe that optimality is proven in the given time limit only for $n < 6$ perturbed labels. 
For $n \geq 6$, our formulation still quantifies a sound certificate. 
This limitation can be attributed to the relatively weaker continuous relaxations induced by larger values of $n$, resulting in more challenging MIQCPs. 
Nevertheless, Figure~\ref{fig:halfmoons_boundaries} (left) shows the ability of the proposed framework to \textit{simultaneously} provide users with worst-case attack patterns and certified model performance. 

\paragraph{Comparison with Incomplete Certification Baselines.}
When the MIQCP solver times out before proving optimality, the final dual bound provides a sound but incomplete poisoning certificate, comparable to existing incomplete methods \cite{lorenz2024bicert,lorenz2024fullcert,sosnin2024certified,sosnin2025abstractgradienttrainingunified}.
In all experimental settings considered, existing incomplete methods yield only trivial guarantees (certified test accuracy of $0\%$), due to the smaller dataset sizes than those considered in these prior works.  
Additionally, our MIQCP uses big-$M$ constants derived from the same relaxation-based procedures, 
optionally tightened via OBBT, 
so the root relaxation is guaranteed to be at least as tight as the corresponding incomplete certificate.  
Thus, even when optimality is not proven, our bounds are no weaker than those provided by existing incomplete approaches under the same threat model.

\section{Conclusion}
\label{sec:conclusion}

In this work we present a comprehensive mixed-integer quadratically constrained programming framework for the exact certification of robustness to training-time data perturbations in gradient-based learning.
By explicitly encoding adversarial data manipulation, training dynamics, and test-time evaluation, our approach provides sound and complete robustness guarantees under general data poisoning threat models.
To improve tractability, we introduce formulation-tightening techniques and primal heuristics to guide the optimization process.

\section{Acknowledgements}

This work was supported by the Turing’s Defence and Security programme through a partnership with the UK government in accordance with the framework agreement between GCHQ \& The Alan Turing Institute.  CT was supported by a BASF/Royal Academy of Engineering Senior Research Fellowship.

\bibliographystyle{splncs04}
\bibliography{references}

@article{gleixner2017three,
  title={Three enhancements for optimization-based bound tightening},
  author={Gleixner, Ambros M and Berthold, Timo and M{\"u}ller, Benjamin and Weltge, Stefan},
  journal={Journal of Global Optimization},
  volume={67},
  number={4},
  pages={731--757},
  year={2017},
  publisher={Springer}
}

@article{efron2004least,
  title={Least angle regression},
  author={Efron, Bradley and Hastie, Trevor and Johnstone, Iain and Tibshirani, Robert},
  year={2004}
}

@article{duong2023dpll,
  title={A dpll (t) framework for verifying deep neural networks},
  author={Duong, Hai and Nguyen, ThanhVu and Dwyer, Matthew},
  journal={arXiv preprint arXiv:2307.10266},
  year={2023}
}

@article{anderson1936species,
  title={The species problem in Iris},
  author={Anderson, Edgar},
  journal={Annals of the Missouri Botanical Garden},
  volume={23},
  number={3},
  pages={457--509},
  year={1936},
  publisher={JSTOR}
}

@article{sosnin2025abstractgradienttrainingunified,
  title={Abstract Gradient Training: A Unified Certification Framework for Data Poisoning, Unlearning, and Differential Privacy},
  author={Sosnin, Philip and Wicker, Matthew and Collyer, Josh and Tsay, Calvin},
  journal={arXiv preprint arXiv:2511.09400},
  year={2025}
}

@inproceedings{dvijotham2018dual,
  title={A Dual Approach to Scalable Verification of Deep Networks.},
  author={Dvijotham, Krishnamurthy and Stanforth, Robert and Gowal, Sven and Mann, Timothy A and Kohli, Pushmeet},
  booktitle={UAI},
  volume={1},
  pages={3},
  year={2018}
}

@inproceedings{rosenfeld2020certified,
  title={Certified robustness to label-flipping attacks via randomized smoothing},
  author={Rosenfeld, Elan and Winston, Ezra and Ravikumar, Pradeep and Kolter, Zico},
  booktitle={International Conference on Machine Learning},
  pages={8230--8241},
  year={2020},
  organization={PMLR}
}

@inproceedings{biggio2014poisoning,
  title={Poisoning behavioral malware clustering},
  author={Biggio, Battista and Rieck, Konrad and Ariu, Davide and Wressnegger, Christian and Corona, Igino and Giacinto, Giorgio and Roli, Fabio},
  booktitle={Proceedings of the 2014 workshop on artificial intelligent and security workshop},
  pages={27--36},
  year={2014}
}

@article{fischetti2018deep,
  title={Deep neural networks and mixed integer linear optimization},
  author={Fischetti, Matteo and Jo, Jason},
  journal={Constraints},
  volume={23},
  number={3},
  pages={296--309},
  year={2018},
  publisher={Springer}
}

@article{tsay2021partition,
  title={Partition-based formulations for mixed-integer optimization of trained {ReLU} neural networks},
  author={Tsay, Calvin and Kronqvist, Jan and Thebelt, Alexander and Misener, Ruth},
  journal={Advances in neural information processing systems},
  volume={34},
  pages={3068--3080},
  year={2021}
}

@article{sabanayagam2024exact,
  title={Exact Certification of (Graph) Neural Networks Against Label Poisoning},
  author={Sabanayagam, Mahalakshmi and Gosch, Lukas and G{\"u}nnemann, Stephan and Ghoshdastidar, Debarghya},
  journal={arXiv preprint arXiv:2412.00537},
  year={2024}
}

@article{sosnin2024certified,
  title={Certified Robustness to Data Poisoning in Gradient-Based Training},
  author={Sosnin, Philip and M{\"u}ller, Mark N and Baader, Maximilian and Tsay, Calvin and Wicker, Matthew},
  journal={Transactions on Machine Learning Research},
  number={2835-8856},
  year={2025}
}

@inproceedings{wicker2024certification,
  title={Certification for Differentially Private Prediction in Gradient-Based Training},
  author={Wicker, Matthew and Sosnin, Philip and Shilov, Igor and Janik, Adrianna and M{\"u}ller, Mark N and de Montjoye, Yves-Alexandre and Weller, Adrian and Tsay, Calvin},
  booktitle={Forty-second International Conference on Machine Learning},
  year={2025}
}

@article{tjeng2017evaluating,
  title={Evaluating robustness of neural networks with mixed integer programming},
  author={Tjeng, Vincent and Xiao, Kai and Tedrake, Russ},
  journal={arXiv preprint arXiv:1711.07356},
  year={2017}
}

@inproceedings{wong2018provable,
  title={Provable defenses against adversarial examples via the convex outer adversarial polytope},
  author={Wong, Eric and Kolter, Zico},
  booktitle={International conference on machine learning},
  pages={5286--5295},
  year={2018},
  organization={PMLR}
}

@inproceedings{weng2018towards,
  title={Towards fast computation of certified robustness for {ReLU}relu networks},
  author={Weng, Lily and Zhang, Huan and Chen, Hongge and Song, Zhao and Hsieh, Cho-Jui and Daniel, Luca and Boning, Duane and Dhillon, Inderjit},
  booktitle={International Conference on Machine Learning},
  pages={5276--5285},
  year={2018},
  organization={PMLR}
}

@article{goodfellow2014explaining,
  title={Explaining and harnessing adversarial examples},
  author={Goodfellow, Ian J and Shlens, Jonathon and Szegedy, Christian},
  journal={arXiv preprint arXiv:1412.6572},
  year={2014}
}

@article{szegedy2013intriguing,
  title={Intriguing properties of neural networks},
  author={Szegedy, Christian and Zaremba, Wojciech and Sutskever, Ilya and Bruna, Joan and Erhan, Dumitru and Goodfellow, Ian and Fergus, Rob},
  journal={arXiv preprint arXiv:1312.6199},
  year={2013}
}

@inproceedings{katz2017reluplex,
  title={Reluplex: An efficient {SMT} solver for verifying deep neural networks},
  author={Katz, Guy and Barrett, Clark and Dill, David L and Julian, Kyle and Kochenderfer, Mykel J},
  booktitle={Computer Aided Verification: 29th International Conference, CAV 2017, Heidelberg, Germany, July 24-28, 2017, Proceedings, Part I 30},
  pages={97--117},
  year={2017},
  organization={Springer}
}

@article{anderson2020strong,
  title={Strong mixed-integer programming formulations for trained neural networks},
  author={Anderson, Ross and Huchette, Joey and Ma, Will and Tjandraatmadja, Christian and Vielma, Juan Pablo},
  journal={Mathematical Programming},
  volume={183},
  number={1},
  pages={3--39},
  year={2020},
  publisher={Springer}
}

@article{muller2022certified,
  title={Certified training: Small boxes are all you need},
  author={M{\"u}ller, Mark Niklas and Eckert, Franziska and Fischer, Marc and Vechev, Martin},
  journal={arXiv preprint arXiv:2210.04871},
  year={2022}
}

@article{gowal2018effectiveness,
  title={On the effectiveness of interval bound propagation for training verifiably robust models},
  author={Gowal, Sven and Dvijotham, Krishnamurthy and Stanforth, Robert and Bunel, Rudy and Qin, Chongli and Uesato, Jonathan and Arandjelovic, Relja and Mann, Timothy and Kohli, Pushmeet},
  journal={arXiv preprint arXiv:1810.12715},
  year={2018}
}

@article{li2020learning,
  title={Learning to detect malicious clients for robust federated learning},
  author={Li, Suyi and Cheng, Yong and Wang, Wei and Liu, Yang and Chen, Tianjian},
  journal={arXiv preprint arXiv:2002.00211},
  year={2020}
}

@article{hong2020effectiveness,
  title={On the effectiveness of mitigating data poisoning attacks with gradient shaping},
  author={Hong, Sanghyun and Chandrasekaran, Varun and Kaya, Yi{\u{g}}itcan and Dumitra{\c{s}}, Tudor and Papernot, Nicolas},
  journal={arXiv preprint arXiv:2002.11497},
  year={2020}
}

@article{wong2018scaling,
  title={Scaling provable adversarial defenses},
  author={Wong, Eric and Schmidt, Frank and Metzen, Jan Hendrik and Kolter, J Zico},
  journal={Advances in Neural Information Processing Systems},
  volume={31},
  year={2018}
}

@inproceedings{wang2022improved,
  title={Improved certified defenses against data poisoning with (deterministic) finite aggregation},
  author={Wang, Wenxiao and Levine, Alexander J and Feizi, Soheil},
  booktitle={International Conference on Machine Learning},
  pages={22769--22783},
  year={2022},
  organization={PMLR}
}

@article{levine2020deep,
  title={Deep partition aggregation: Provable defense against general poisoning attacks},
  author={Levine, Alexander and Feizi, Soheil},
  journal={arXiv preprint arXiv:2006.14768},
  year={2020}
}

@inproceedings{rezaei2023run,
  title={Run-off election: Improved provable defense against data poisoning attacks},
  author={Rezaei, Keivan and Banihashem, Kiarash and Chegini, Atoosa and Feizi, Soheil},
  booktitle={International Conference on Machine Learning},
  pages={29030--29050},
  year={2023},
  organization={PMLR}
}

@inproceedings{munoz2017towards,
  title={Towards poisoning of deep learning algorithms with back-gradient optimization},
  author={Mu{\~n}oz-Gonz{\'a}lez, Luis and Biggio, Battista and Demontis, Ambra and Paudice, Andrea and Wongrassamee, Vasin and Lupu, Emil C and Roli, Fabio},
  booktitle={Proceedings of the 10th ACM workshop on artificial intelligence and security},
  pages={27--38},
  year={2017}
}

@article{gu2017badnets,
  title={Badnets: Identifying vulnerabilities in the machine learning model supply chain},
  author={Gu, Tianyu and Dolan-Gavitt, Brendan and Garg, Siddharth},
  journal={arXiv preprint arXiv:1708.06733},
  year={2017}
}

@article{chen2017targeted,
  title={Targeted backdoor attacks on deep learning systems using data poisoning},
  author={Chen, Xinyun and Liu, Chang and Li, Bo and Lu, Kimberly and Song, Dawn},
  journal={arXiv preprint arXiv:1712.05526},
  year={2017}
}

@inproceedings{newsome2006paragraph,
  title={Paragraph: Thwarting signature learning by training maliciously},
  author={Newsome, James and Karp, Brad and Song, Dawn},
  booktitle={Recent Advances in Intrusion Detection: 9th International Symposium, RAID 2006 Hamburg, Germany, September 20-22, 2006 Proceedings 9},
  pages={81--105},
  year={2006},
  organization={Springer}
}

@inproceedings{biggio2018wild,
  title={Wild patterns: Ten years after the rise of adversarial machine learning},
  author={Biggio, Battista and Roli, Fabio},
  booktitle={Proceedings of the 2018 ACM SIGSAC Conference on Computer and Communications Security},
  pages={2154--2156},
  year={2018}
}

@article{tian2022comprehensive,
  title={A comprehensive survey on poisoning attacks and countermeasures in machine learning},
  author={Tian, Zhiyi and Cui, Lei and Liang, Jie and Yu, Shui},
  journal={ACM Computing Surveys},
  volume={55},
  number={8},
  pages={1--35},
  year={2022},
  publisher={ACM New York, NY}
}

@inproceedings{han2022physical,
  title={Physical backdoor attacks to lane detection systems in autonomous driving},
  author={Han, Xingshuo and Xu, Guowen and Zhou, Yuan and Yang, Xuehuan and Li, Jiwei and Zhang, Tianwei},
  booktitle={Proceedings of the 30th ACM International Conference on Multimedia},
  pages={2957--2968},
  year={2022}
}

@inproceedings{zhu2019transferable,
  title={Transferable clean-label poisoning attacks on deep neural nets},
  author={Zhu, Chen and Huang, W Ronny and Li, Hengduo and Taylor, Gavin and Studer, Christoph and Goldstein, Tom},
  booktitle={International conference on machine learning},
  pages={7614--7623},
  year={2019},
  organization={PMLR}
}

@inproceedings{yang2017fake,
  title={Fake Co-visitation Injection Attacks to Recommender Systems.},
  author={Yang, Guolei and Gong, Neil Zhenqiang and Cai, Ying},
  booktitle={NDSS},
  year={2017}
}

@inproceedings{carlini2023poisoning,
  title={Poisoning web-scale training datasets is practical},
  author={Carlini, Nicholas and Jagielski, Matthew and Choquette-Choo, Christopher A and Paleka, Daniel and Pearce, Will and Anderson, Hyrum and Terzis, Andreas and Thomas, Kurt and Tram{\`e}r, Florian},
  booktitle={2024 IEEE Symposium on Security and Privacy (SP)},
  pages={407--425},
  year={2024},
  organization={IEEE}
}

@article{xie2022uncovering,
  title={Uncovering the Connection Between Differential Privacy and Certified Robustness of Federated Learning against Poisoning Attacks},
  author={Xie, Chulin and Long, Yunhui and Chen, Pin-Yu and Li, Bo},
  journal={arXiv preprint arXiv:2209.04030},
  year={2022}
}

@article{steinhardt2017certified,
  title={Certified defenses for data poisoning attacks},
  author={Steinhardt, Jacob and Koh, Pang Wei W and Liang, Percy S},
  journal={Advances in neural information processing systems},
  volume={30},
  year={2017}
}

@article{singh2019,
author = {Singh, Gagandeep and Gehr, Timon and P\"{u}schel, Markus and Vechev, Martin},
title = {An abstract domain for certifying neural networks},
year = {2019},
issue_date = {January 2019},
publisher = {Association for Computing Machinery},
address = {New York, NY, USA},
volume = {3},
number = {POPL},
doi = {10.1145/3290354},
journal = {Proc. ACM Program. Lang.},
month = {jan},
articleno = {41},
numpages = {30},
}

@article{zhang2018efficient,
  title={Efficient neural network robustness certification with general activation functions},
  author={Zhang, Huan and Weng, Tsui-Wei and Chen, Pin-Yu and Hsieh, Cho-Jui and Daniel, Luca},
  journal={Advances in neural information processing systems},
  volume={31},
  year={2018}
}

@article{huchette2023deep,
  title={When deep learning meets polyhedral theory: A survey},
  author={Huchette, Joey and Mu{\~n}oz, Gonzalo and Serra, Thiago and Tsay, Calvin},
  journal={arXiv preprint arXiv:2305.00241},
  year={2023}
}

@inproceedings{sosnin2024scaling,
  title={Scaling mixed-integer programming for certification of neural network controllers using bounds tightening},
  author={Sosnin, Philip and Tsay, Calvin},
  booktitle={2024 IEEE 63rd Conference on Decision and Control (CDC)},
  pages={1645--1650},
  year={2024},
  organization={IEEE}
}

@article{bunel2018unified,
  title={A unified view of piecewise linear neural network verification},
  author={Bunel, Rudy R and Turkaslan, Ilker and Torr, Philip and Kohli, Pushmeet and Mudigonda, Pawan K},
  journal={Advances in Neural Information Processing Systems},
  volume={31},
  year={2018}
}

@inproceedings{ehlers_formal_2017,
  title={Formal verification of piece-wise linear feed-forward neural networks},
  author={Ehlers, Ruediger},
  booktitle={International symposium on automated technology for verification and analysis},
  pages={269--286},
  year={2017},
  organization={Springer}
}

@article{lorenz2024bicert,
  title={BiCert: A Bilinear Mixed Integer Programming Formulation for Precise Certified Bounds Against Data Poisoning Attacks},
  author={Lorenz, Tobias and Kwiatkowska, Marta and Fritz, Mario},
  journal={arXiv preprint arXiv:2412.10186},
  year={2024}
}

@inproceedings{lorenz2024fullcert,
  title={Fullcert: Deterministic end-to-end certification for training and inference of neural networks},
  author={Lorenz, Tobias and Kwiatkowska, Marta and Fritz, Mario},
  booktitle={DAGM German Conference on Pattern Recognition},
  pages={71--85},
  year={2024},
  organization={Springer}
}

@inproceedings{bose2025keeping,
  title={Keeping up with dynamic attackers: Certifying robustness to adaptive online data poisoning},
  author={Bose, Avinandan and Lessard, Laurent and Fazel, Maryam and Dvijotham, Krishnamurthy Dj},
  booktitle={International Conference on Artificial Intelligence and Statistics},
  pages={4438--4446},
  year={2025},
  organization={PMLR}
}

\end{document}